\documentclass[final]{cvpr}

\usepackage{times}
\usepackage{epsfig}
\usepackage{graphicx}
\usepackage{amsmath}
\usepackage{amssymb}

\usepackage[pagebackref=true,breaklinks=true,colorlinks,bookmarks=false]{hyperref}

\def\cvprPaperID{****} 
\def\confYear{CVPR 2021}

\begin{document}

\title{Insights into Data through Model Behaviour: An Explainability-driven Strategy for Data Auditing for Responsible Computer Vision Applications}

\author{Alexander Wong$^{1,2,3,*}$, Adam Dorfman$^{3}$, Paul McInnis$^{3}$, and Hayden Gunraj$^{1}$\\
$^1$Department of Systems Design Engineering, University of Waterloo\\
$^2$Waterloo Artificial Intelligence Institute\\ $^3$DarwinAI Corp.\\
}

\maketitle

\begin{abstract}
In this study, we take a departure and explore an explainability-driven strategy to data auditing, where actionable insights into the data at hand are discovered through the eyes of quantitative explainability on the behaviour of a dummy model prototype when exposed to data.  We demonstrate this strategy by auditing two popular medical benchmark datasets, and discover hidden data quality issues that lead deep learning models to make predictions for the wrong reasons.  The actionable insights gained from this explainability driven data auditing strategy is then leveraged to address the discovered issues to enable the creation of high-performing deep learning models with appropriate prediction behaviour.  The hope is that such an explainability-driven strategy can be complimentary to data-driven strategies to facilitate for more responsible development of machine learning algorithms for computer vision applications.

\end{abstract}
\section{Introduction}
\label{introduction}

The rise of open source benchmark datasets~\cite{russakovsky2015imagenet,lin2015microsoft,VOC,Wang_2017,medmnist,covidnet,alex2020covidnets} has led to significant progress in machine learning for computer vision.  A common assumption made when leveraging benchmark datasets is that they are curated in a way that is free of data quality issues.  Therefore, such datasets are often used 'as is' and sight unseen in practice to train new models, relying on scalar performance metrics to judge a model's efficacy.  However, data auditing in recent studies~\cite{dulhanty2019auditing,prabhu2020large,wang2019balanced} have unveiled hidden biases and data quality issues in well-known benchmark datasets, which can negatively affect real-world performance of models trained on such datasets.

Most data auditing strategies consider only data characteristics and not model behaviour, and thus such strategies are subjective, based largely on human intuition, and can leave hidden issues that affect model behaviour negatively.  In this study, we take a different approach and explore an explainability-driven strategy to data auditing, where actionable insights into data are discovered through the eyes of explainability based on a model's behaviour.

This paper is organized as follows.  Section~\label{Methodology} describes the underlying methodology behind explainability-driven strategy to data for the development of responsible computer vision applications auditing.  Section 2 describes the experiments conducted on two popular medical benchmark datasets.  Section 3 presents the results in terms of the hidden data quality issues that were discovered during the auditing process, as well as actionable insights and steps taken as a result of such insights.  Finally, conclusions are drawn in Section 4.

{
\begin{figure*}[t]
\centering
    \includegraphics[width=1\textwidth]{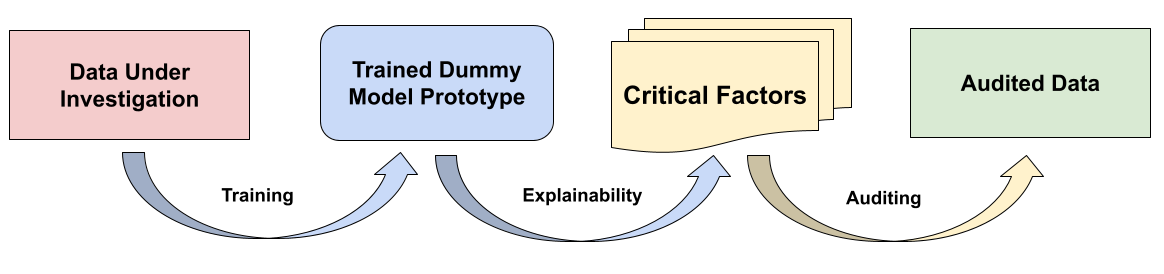}
    \caption{Overview of the explainability-driven data auditing workflow. A dummy model prototype is trained using the data under investigation.  A quantitative explainability technique is used to identify critical factors in the data that drives the prediction behaviour of the prototype.  The critical factors are audited to discover hidden data quality issues.}
    \label{fig:workflow}
\end{figure*}
}

\section{Methodology}
\label{Methodology}

In this study, we aim to explore the efficacy of an explainability-driven strategy to data auditing for the development of responsible computer vision applications, which is a conceptual departure from the direction taken by existing data-driven strategies in research literature.  More specifically,  explainability-driven data auditing is conducted as follows (see Figure~\ref{fig:workflow}).  First, a dummy model prototype is constructed and trained with the data under investigation. Second, the data is fed back into the trained dummy model prototype and a quantitative explainability technique is leveraged to identify the critical factors driving the behaviour of the prototype across the data.  Third, the identified critical factors are studied to discover hidden data quality issues.  An unique aspect of this explainability-driven strategy to data auditing is that hidden data quality issues are discovered based on prediction behaviour through the eyes of explainability, and as such has the potential to compliment data-driven strategies to uncover hidden issues not identified based on considering just data characteristics alone.

\section{Experiments}

We demonstrate this strategy by auditing two popular benchmark datasets (OSIC Pulmonary Fibrosis Progression dataset~\cite{OSIC} and CNCB COVID-19 CT dataset~\cite{cncb}) on dummy deep CNN regression and classification model prototypes, respectively. For explainability we leverage GSInquire~\cite{gsinquire}, which was demonstrated to better reflect a model's decision-making process when compared to state-of-the-art approaches, and one of the only approaches that can be used on deep learning regression models.  In particular, the OSIC dataset is part of a popular Kaggle challenge, with the winning solution~\cite{winner} using the data largely 'as is' without considering the CT modality used. These are good use cases given the importance of responsible computer vision in healthcare.

\section{Results and Discussion}

In this section, we will discuss the hidden data quality issues that were discovered using the proposed explainability-driven strategy for data auditing, as well as the steps taken to address these issues based on the actionable insights gained from the data auditing strategy.

{
\begin{figure}[b]
\centering
    \includegraphics[width=0.5\textwidth]{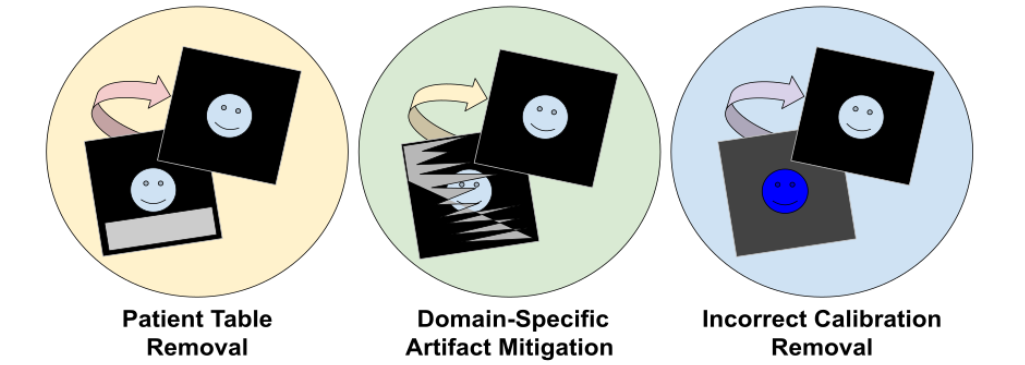}
    \caption{Actionable insights gained from the explainability-driven data auditing process can be used to address discovered data quality issues.}
    \label{fig:insights}
\end{figure}
}

\subsection{Hidden Data Quality Issues Discovery}
The explainability-driven data auditing led to the discovery of several hidden data quality issues that caused the dummy model prototypes to make predictions for the wrong reasons, even if performance is high based on scalar metrics.  These include: 1) \textbf{incorrect calibration metadata} led to data dynamic range being erroneously used by the model prototype to make predictions, 2) \textbf{synthetic padding} (Fig.~\ref{fig:issues}a) introduced during data curation being used to erroneously guide predictions, 3) \textbf{circular artifacts} (Fig.~\ref{fig:issues}b) being used by the model to erroneously guide predictions, and 4) \textbf{patient tables} (Fig.~\ref{fig:issues}c) being used by the model to make predictions.

{
\begin{figure*}[t]
\centering
    \includegraphics[width=0.8\textwidth]{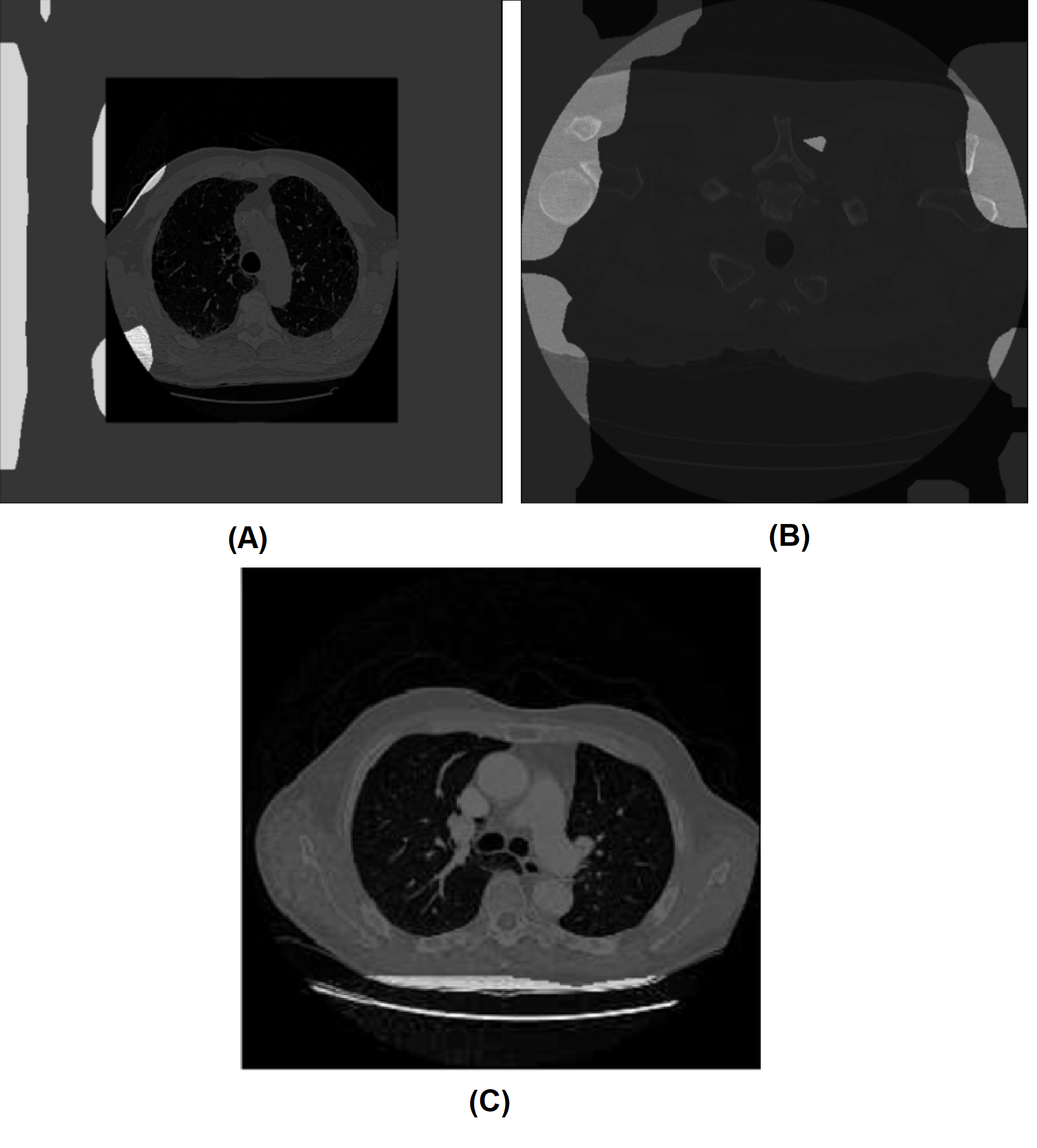}
    \caption{Example discovered data quality issues. Highlighted are critical factors used by model prototypes found via explainability. (A) synthetic padding, (B) circular artifacts, (C) patient tables.}
    \label{fig:issues}
\end{figure*}
}

\subsection{Actionable Insights}
The discovered data quality issues led to the following actionable insights: 1) incorrect calibration data removal, 2-3) domain-specific artifact mitigation, and 4) automatic table removal (see Figure~\ref{fig:insights}).  By taking the above actions on the data set to address the discovered data quality issues uncovered via the aforementioned explainability-driven strategy for data auditing, the resulting deep learning models not only achieved significantly higher performance~\cite{wong2021fibrosisnet,covidnetct}, but also led to models that made predictions based on the right visual cues.

For example, in the case of the OSIC Pulmonary Fibrosis Progression dataset, addressing the discovered data quality issues led to the creation of a deep CNN regression model~\cite{wong2021fibrosisnet} with state-of-the-art performance above the winning solutions in the OSIC Kaggle Challenge~\cite{winner} that learned to leverage relevant visual anomalies such as honeycombing in the lungs (see Fig.~\ref{fig:fibrosis} for example CT images from the OSIC Pulmonary Fibrosis Progression dataset and corresponding identified critical factors).  In the case of CNCB COVID-19 CT dataset, addressing the discovered data quality issues led to the creation of deep CNN classification models~\cite{covidnetct,gunraj2021covidnet} with state-of-the-art performance (exceeding 98\% accuracy) that learned to leverage relevant visual anomalies in the lungs such as ground-glass opacities and bilateral bilateral patchy opacities.

{
\begin{figure*}[h]
\centering
    \includegraphics[width=\textwidth]{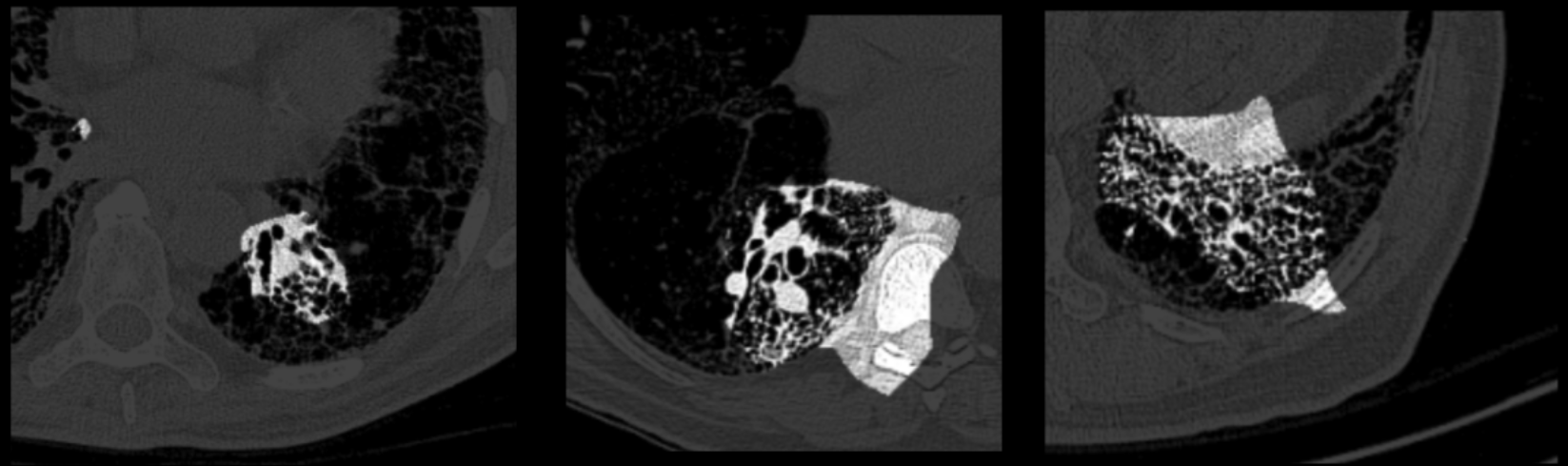}
    \caption{Example CT images from the OSIC Pulmonary Fibrosis Progression dataset, with the highlighted areas corresponding to the critical factors used by a state-of-the-art deep CNN regression model for predicting fibrosis progression, as identified via quantitative explainability.  It can be observed that, by addressing the discovered data quality issues identified via explainability-driven data auditing, the model is able to learn to leverage relevant visual anomalies such as honeycombing in the lungs.}
    \label{fig:fibrosis}
\end{figure*}
}

\section{Conclusions}
In this study, an explainability-driven strategy for data auditing was explored and conducted on two different popular medical benchmark datasets.  The proposed data auditing strategy led to the discovery of critical hidden data quality issues that led to incorrect prediction behaviour of deep learning models, and the actionable insights gained was leveraged to addressing these issues and enable the creation of deep learning models that not only had higher performance but also made predictions based on the right reasons.  The hope is this explainability-driven strategy can compliment data-driven strategies to facilitate for more responsible machine learning-driven computer vision development.

{\small
\bibliographystyle{ieee_fullname}
\bibliography{egbib}
}

\end{document}